\documentclass[conference]{IEEEtran}
\IEEEoverridecommandlockouts

\usepackage{cite}
\usepackage{amsmath,amssymb,amsfonts}
\usepackage{algorithmic}
\usepackage{dsfont}
\usepackage[ruled, vlined]{algorithm2e}
\usepackage{hyperref}
\usepackage{graphicx}
\usepackage[caption=false]{subfig}
\usepackage{textcomp}
\usepackage[dvipsnames]{xcolor}
\def\BibTeX{{\rm B\kern-.05em{\sc i\kern-.025em b}\kern-.08em
    T\kern-.1667em\lower.7ex\hbox{E}\kern-.125emX}}
\newcommand{\e}{\mathrm{e}}
\newcommand{\dd}{\mathrm{d}}

\newcommand{\sg}{\mathbf{s}}
\newcommand{\ub}{\mathbf{u}}
\newcommand{\vb}{\mathbf{v}}
\newcommand{\gb}{\mathbf{g}}

\begin{document}

\title{Disentangling Modes and Interference in the Spectrogram of Multicomponent Signals\\
\thanks{Email of the corresponding author: kevin.polisano@univ-grenoble-alpes.fr.}
}

\author{\IEEEauthorblockN{Kévin Polisano\IEEEauthorrefmark{1}, Sylvain Meignen\IEEEauthorrefmark{1}, Nils Laurent\IEEEauthorrefmark{2}, Hubert Leterme\IEEEauthorrefmark{3}} \\
\IEEEauthorblockA{\IEEEauthorrefmark{1} Univ. Grenoble Alpes, CNRS, F-38000 Grenoble, France \\ \IEEEauthorrefmark{2} Ens de Lyon, CNRS, Laboratoire de Physique, F-69342 Lyon, France \\ \IEEEauthorrefmark{3} Université Caen Normandie, ENSICAEN, CNRS, Normandie Univ, GREYC UMR 6072, F-14000 Caen, France}
}


\maketitle

\begin{abstract}
In this paper, we investigate how the spectrogram of multicomponent signals can be decomposed into a mode part and an interference part. We explore two approaches: (i) a variational method inspired by texture-geometry decomposition in image processing, and (ii) a supervised learning approach using a U-Net architecture, trained on a dataset encompassing diverse interference patterns and noise conditions. Once the interference component is identified, we explain how it enables us 
to define a criterion to locally adapt the window length used in the definition of the spectrogram, for the sake of improving ridge detection in the presence of close modes. 
Numerical experiments illustrate the advantages and limitations of both approaches for spectrogram decomposition, highlighting their potential for enhancing time-frequency analysis in the presence of strong interference.
\end{abstract}


\section{Introduction}

\IEEEPARstart{N}{on}-stationary signals such as audio (music, speech, bird songs) \cite{gribonval2003harmonic}, electrocardiogram \cite{Herry2017} or thoracic and abdominal movement signals \cite{Lin2016} can be modelled as a superimposition of amplitude and frequency-modulated (AM/FM) modes, referred to as \emph{multicomponent signal} (MCS), and 
defined as 
\begin{equation}
\label{def:MCS}
f(t) = \sum_{p=1}^P f_p(t), \textrm{ with } f_p(t) = A_p(t)\e^{2i\pi \phi_p(t)},
\end{equation}
where the \emph{instantaneous amplitudes} (IAs) $A_p(t)$ and the \emph{instantaneous frequencies} (IFs) $\phi_p'(t)$ are set to be positive.
This representation is associated with the ideal \emph{time-frequency} (TF) signature defined as: 
\begin{equation}
\label{def:ITF}
IT_f(t, \eta) = \sum_{p=1}^P A_p(t) \delta (\eta - \phi_p'(t)),
\end{equation}
where $\delta$ is the Dirac distribution.
To estimate $IT_f$, the \emph{short-time Fourier transform} (STFT) is commonly used
\begin{eqnarray}
	\label{def:STFT}
	V_f^h(t, \eta) = \int_{\mathbb{R}} f(x) h(x-t) \e^{-i 2\pi \eta (x-t)} \dd x,
\end{eqnarray}
where $h$ is a real-valued window. The squared modulus of the STFT, known as the \emph{spectrogram}, is given by 
$S^{h}_f := |V_f^h|^2$. The ideal TF signature associated with the spectrogram is the 
same as $\eqref{def:ITF}$ replacing the amplitudes with their squared values.

Traditionally, IFs are estimated on the spectrogram by considering \emph{ridge detection} (RD), which involves identifying and connecting local maxima along the frequency axis over time \cite{delprat1997global}. In connection with IF estimation, significant efforts have been made to enhance the readability of TF representations, including reassignment methods \cite{auger2013time} and the \textit{Fourier-based synchrosqueezing transform} (FSST) \cite{oberlin2015second}. 
However, all these techniques become irrelevant in the presence of strong interference in the TF plane. 
Indeed, when two or more modes are close or intersect in the TF plane, reassignment operators become unreliable, while a mode may no longer be associated with a single spectrogram ridge. 
The resulting phase mixing leads to constructive or destructive interference, creating complex patterns that are difficult to analyze \cite{meignen2022one}.  More precisely, the spectrogram of a MCS $f$, as defined in \eqref{def:MCS}, is the sum of the spectrograms of each mode  plus undesirable extra cross-terms responsible of the interference:
\begin{equation}
\label{eq:def_spec}
|V_f^h|^2 = \sum_{p=1}^P |V_{f_p}^h|^2 +2\sum_{p=1}^P \sum_{\substack{q=1\\ q\neq p}}^P \mathrm{Re}(V_{f_p}^h{V_{f_q}^h}^{\ast})\; .
\end{equation}
Thus, to have an insight into where the interference term can be 
neglected is essential for IF estimation for instance.
In this regard, several recent studies have aimed at localizing interference in the TF plane using techniques such as hyperbolic spline fitting \cite{bruni2025multicomponent} or neural 
network \cite{bruniSupervised2023}. The latter work 
primarily focused on segmenting interference regions using a convolutional neural network (CNN)-based classification approach. However, the dataset is limited to linear chirps, does not account for complex interference patterns that may arise, and relies on a single window choice. Moreover, this study neither addresses the separation of interference from the rest of the signal, nor proposes strategies for IF estimation in regions affected by interference.
We shall mention that as the spectrogram is constrained by the choice of a window, the Wigner-Ville distribution (WVD) offers an alternative for IF estimation. For that purpose, one needs to separate
the interference part from the rest of the signal as proposed in \cite{zhangCrosstermfree2022a}. However, in the spectrogram representation interference are mitigated by windowing, but, 
in the WVD representation of a MCS, interference between any two modes remain, leading 
to highly complex patterns when more than two modes are present (see \emph{e.g} \cite[Fig. 12]{zhangCrosstermfree2022a}).

Our goal in the present paper is to stick to the spectrogram representation and investigate how it can be decomposed into a mode part and an interference part. 
To this end, we propose to consider two different paths. The first one consists in revisiting variational approaches involving proximal techniques that were initially proposed to separate texture from 
geometry in images \cite{aujol2006structure}. In our context, we interpret the “geometrical” part of the spectrogram, viewed as an image, as the so-called  "modes part", while the “texture” part corresponds to interference in the spectrogram. 
The second approach we consider relies on supervised learning, employing a U-Net architecture \cite{ronneberger2015u} trained on a dataset designed to capture diverse interference scenarios. Our dataset also incorporates noise to enhance robustness.
Once the interference component has been extracted from the spectrogram, we leverage this information to refine RD using an adaptive optimal window selection for improved spectrogram-based IF estimation.

The paper is organized as follows. In Sections \ref{sec:first_model} and \ref{sec:second_model}, we 
detail the two different techniques we propose to 
separate the mode parts from the interference part 
in the spectrogram. Section \ref{sec:rres_discuss} presents numerical experiments illustrating the advantages and limitations of the proposed methods, along with insights into how interference extraction can benefit ridge estimation.

\section{First Model Based on a Variational Approach}\label{sec:first_model}

\subsection{Oscillating Interference Patterns on Toy Models}\label{sec:toy}

In the case of two parallel harmonics, a simple analytical expression for the spectrogram can be derived. Considering the sum of two harmonics, $f(t) = A_1\e^{i 2\pi f_1 t} + A_2\e^{i 2\pi f_2 t}$ and computing its STFT with a Gaussian window $h_\sigma(x)=\frac{1}{\sigma} e^{-\pi\frac{x^2}{\sigma^2}}$, we obtain
\begin{equation}
\label{eq:spect}
\begin{aligned}
		& |V_{f}^{h_\sigma}(t, \eta)|^{2}= \overbrace{A_1^{2}e^{-2\pi \sigma^{2}(\eta-f_{1})^{2}} + A_2^2e^{-2\pi \sigma^{2}(\eta-f_{2})^{2}}}^{\text{Mode part}}\\
		&+ \underbrace{2A_1A_2e^{-\pi \sigma^{2}\big((\eta-f_{1})^{2} + (\eta-f_{2})^{2}\big)} \cos(2\pi(f_{2}-f_{1})t)}_{\text{Interference part}},
\end{aligned}
\end{equation}
where the so-called ``Mode part'' (first two terms) represents the individual modes, while the ``Interference part'' (last term) corresponds to the interference pattern. This interference strongly depends on the window length parameter $\sigma$ \cite{meignen2022one}. It can be observed that the interference term oscillates along the time axis with a frequency $\delta f := |f_2-f_1|$ along the time axis, while its amplitude depends only on frequency.

\begin{figure}
    \includegraphics[width=0.5\textwidth]{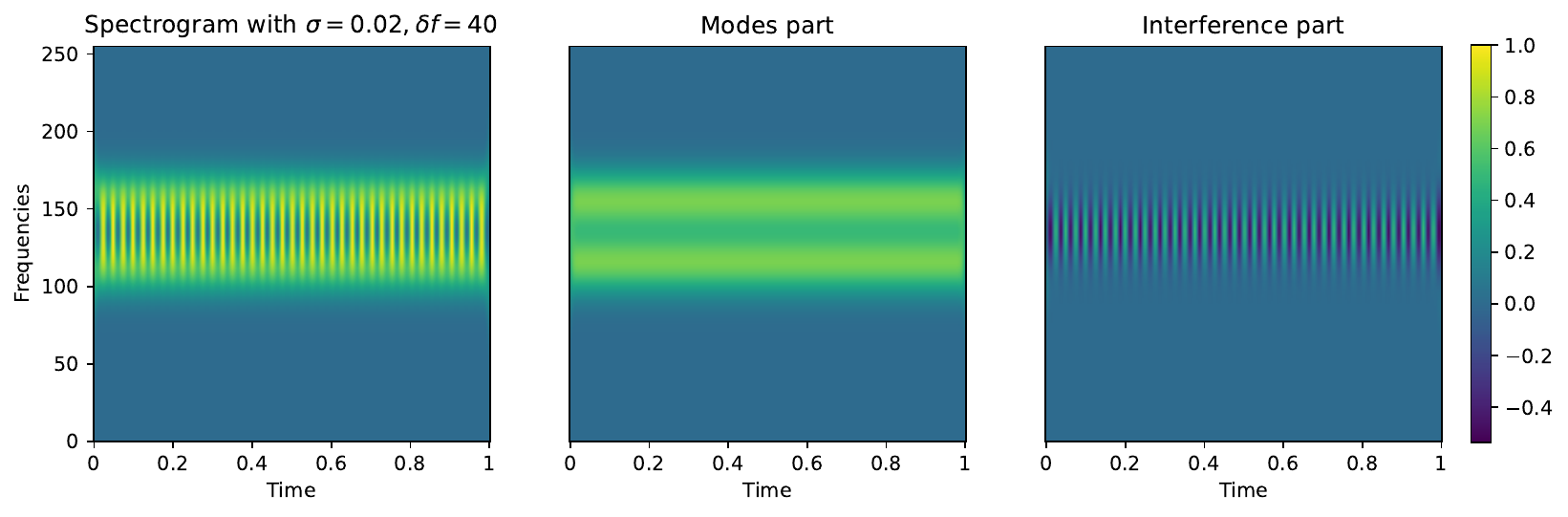}
    \caption{Spectrogram of two harmonics \eqref{eq:spect} separated by $ \delta f = 40Hz$ (left), along with its decomposition into a ``Mode part'' (middle) and an ``Interference part'' (right).}
    \label{fig:harmonics}
\end{figure}

As an example, we compute the spectrogram of two harmonics separated by $ \delta f = 40Hz$. The first harmonic is set at 115Hz, while the second at 155Hz. We consider a signal of length $N=256$ over the interval $[0,1]$ with $N$ frequency-bins and a window size of $ \sigma = 0.01$. The resulting spectrogram, along with the decomposition in ``Mode part'' and ``Interference part'', are shown in Fig.~\ref{fig:harmonics}.
Note that, one can 
obtain analytically the same kind of oscillating pattern for the interference term in the case of two parallel linear chirps \cite{meignen2022one}. 

\subsection{TV-minimization using the $G$-norm}
\label{sec:TVminimization}
A spectrogram containing mode parts and interference part can be viewed as a geometric part (mode parts) overlaid with 
a texture part (interference part). Inspired by image decomposition algorithms that 
separate an image into texture and cartoon part \cite{aujolImageDecompositionBounded2005}, we propose to transpose this approach to the 
case of spectrograms containing interfering modes. In  \cite{aujolImageDecompositionBounded2005}, 
the geometric part is modeled as a function of bounded variations, denoted by $u$, while the texture, denoted by $v$, belongs to a space of oscillatory functions. The spectrogram, denoted by $s$, 
should then be decomposed as $s \approx u+v$. 
For the sake of clarity, we briefly summarize this approach. Assuming that the discrete spectrogram $\sg$ is represented as an $N \times N$ image, we define $\mathcal{X}= \mathbb{R}^{N \times N}$ and $\mathcal{Y} = \mathcal{X} \times \mathcal{X}$.  
 The discrete total variation for any $\ub \in \mathcal{X}$ is given by   
\begin{eqnarray}
J(\ub) =\sum_{\begin{array}{c}1 \leq i\leq N\\
                              1 \leq j \leq N
              \end{array}}                 
            \|(\nabla \ub)_{i,j}\|_2,  
\end{eqnarray}
where the subscript $2$ denotes the Euclidian norm on $\cal Y$. 
To model the oscillations, the subspace
\begin{eqnarray}
\mathcal{G}  = \left \{ \vb \in \mathcal{X}, \exists \gb \in \mathcal{Y} \ s.t \  \vb = \mathrm{div}(\gb) \right \},
\end{eqnarray}
is introduced, where $\mathrm{div}$ denotes the discrete divergence. 
This space can be equipped with the norm: 
\begin{eqnarray}
\|\vb\|_{\mathcal{G}} = \inf\limits_{\gb \in \mathcal{Y}, \vb = \mathrm{div} (\gb)} 
\left \{ \|\gb\|_\infty = \max_{i,j} \|g_{i,j}\|_2 \right \}.  
\end{eqnarray}
$\mathcal{G}_{\mu}$ is then defined as the set of elements in $\mathcal{G}$ 
with norm smaller than $\mu$. The decomposition of the spectrogram 
image $\sg$ into a bounded variation part $\ub$ and a texture part 
$\vb$ is performed by solving the minimization problem  \cite{aujol2005dual}:
\begin{eqnarray}
\label{eq:defmin}
\mathop{\rm argmin}_{(\ub,\vb) \in \mathcal{X} \times \mathcal{G}_\mu}  \frac{1}{2}  \|\sg-\ub-\vb\|_2^2 + \lambda J(\ub),
\end{eqnarray}
the subscript $2$ corresponding to the Euclidian norm on $\mathcal{X}$.   
\subsection{Implementation with Alternate Projections}
\label{sec:algorithm}
Fixing $\ub$, one minimizes \eqref{eq:defmin} by considering: 
\begin{equation}
\label{eq:models:aujolStep1}
\inf\limits_{\vb \in \mathcal{G}_\mu} \| \sg - \ub - \vb \|_2^2.
\end{equation}
Fixing $\vb$, one minimizes \eqref{eq:defmin} as:  
\begin{equation}
\label{eq:models:aujolStep2}
\mathop{\rm argmin}_{\ub \in \mathcal{X}} J(\ub) + \frac{1}{2\lambda} \|\sg - \ub - \vb\|_2^2.
\end{equation}
Eq. \eqref{eq:models:aujolStep1} is equivalent to projecting $\sg - \ub - \vb$ 
onto $\mathcal{G}_\mu$ which is performed using Chambolle's algorithm \cite{chambolle2004algorithm}, 
while the solution to \eqref{eq:models:aujolStep2} is  
$\ub = \sg - \vb - \pi_{\mathcal{G}_\mu}(\sg - \vb)$ (where $\pi$ denotes the projection)\cite{chambolle2004algorithm}. 
Such a decomposition is summarized in Algorithm \ref{Algo1}.
\begin{algorithm}[ht]
\caption{Image separation algorithm\cite{aujol2005dual}}
\label{Algo1}
\begin{algorithmic}[1]
 \STATE \textbf{Initialization: }$\ub_0 = \sg$, $\vb_0 = \vb_1 = \ub_1 = 0, \epsilon$
 \STATE $k = 0$
 \WHILE {$\max\left(\|\ub_{k+1}-\ub_k\|_{\infty}, \|\vb_{k+1}-\vb_k\|_{\infty} \right) > \epsilon$}
  \STATE $\vb_{k+1} = \pi_{\mathcal{G}_\mu}(\sg-\ub_k)$
  \STATE $\ub_{k+1} = \sg - \vb_{k+1} - \pi_{\mathcal{G}_\lambda}(\sg-\vb_{k+1})$
 \ENDWHILE
\end{algorithmic}
\end{algorithm}

Note that with such an approach, the noise, if any, is captured in the texture part of the image. 
It is worth noting here that developments on textured images denoising were proposed based on soft-thresholding, 
assuming the added noise is Gaussian white (see \cite{aujol2005dual}, for instance).   
However, in the case of spectrograms of noisy signals, the noise is filtered and cannot be removed by soft-thresholding as proposed in \cite{aujol2005dual}. 

\section{Second Model Based on Deep Learning}
\label{sec:second_model}

\subsection{A Supervised Approach Based on U-Net}
If the MCS consists of two modes corrupted by some noise, denoted as $\tilde f = f_1+f_2+\epsilon$, its spectrogram reads:
\begin{IEEEeqnarray}{lCr}
    |V_{\tilde f}^h
    |^2 &=& |V_{f_1}^h|^2 + |V_{f_2}^h|^2 \label{eq:t1} \\ &+& |V_{\epsilon}^h|^2 \label{eq:t2} \\ &+& 2\mathrm{Re}(V_{f_1}^h{V_{f_2}^h}^{\ast}) \label{eq:t3} \\ &+& 
    2\mathrm{Re}(V_{f_1}^h{V_{\epsilon}^h}^{\ast}) + 2\mathrm{Re}(V_{f_2}^h{V_{\epsilon}^h}^{\ast}), \label{eq:t4}
\end{IEEEeqnarray}
where \eqref{eq:t1} corresponds to the ideal spectrogram of the two modes (free of noise and interference),  \eqref{eq:t2} to the noise spectrogram,  \eqref{eq:t3} to the interference between modes,  
and \eqref{eq:t4} to the interference between the noise and the modes.

\begin{figure}
    \includegraphics[width=\linewidth]{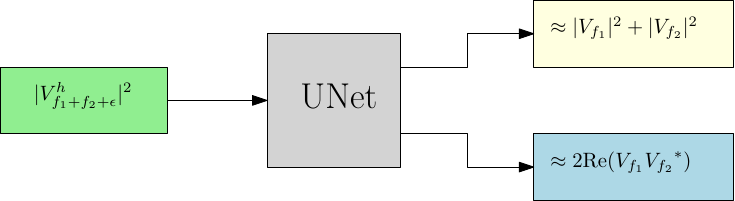}
    \caption{Illustration of the U-Net decomposition of the spectrogram \eqref{eq:t1}-\eqref{eq:t3}.}
\label{Fig:U-NET}
\end{figure}

The objective is to train a neural network that, given a noisy spectrogram $|V_{\tilde f}^h|^2$, can extract both the ideal spectrogram \eqref{eq:t1} and the interference \eqref{eq:t3}, while discarding 
both the noise spectrogram \eqref{eq:t2} and the interference components due to noise \eqref{eq:t4}. A U-Net model \cite{ronneberger2015u} is used as the backbone due to its ability to capture both local and global features through skip connections. An overview of the U-Net decomposition is shown in Fig. \ref{Fig:U-NET}.

\begin{figure*}
    \subfloat[]{ \includegraphics[ width=0.145\textwidth]{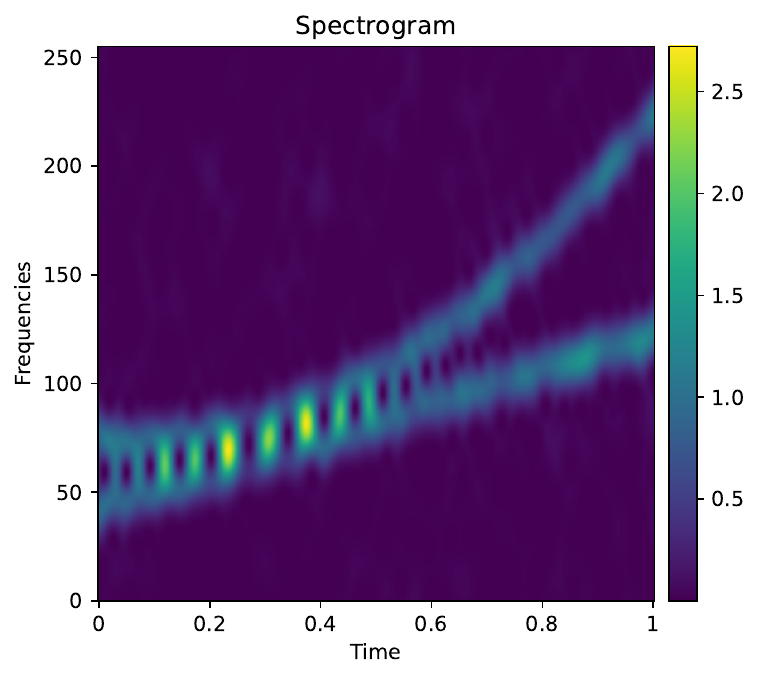}\label{figa:spec_hyperb}
    }
    \subfloat[]{ \includegraphics[ width=0.4\textwidth]{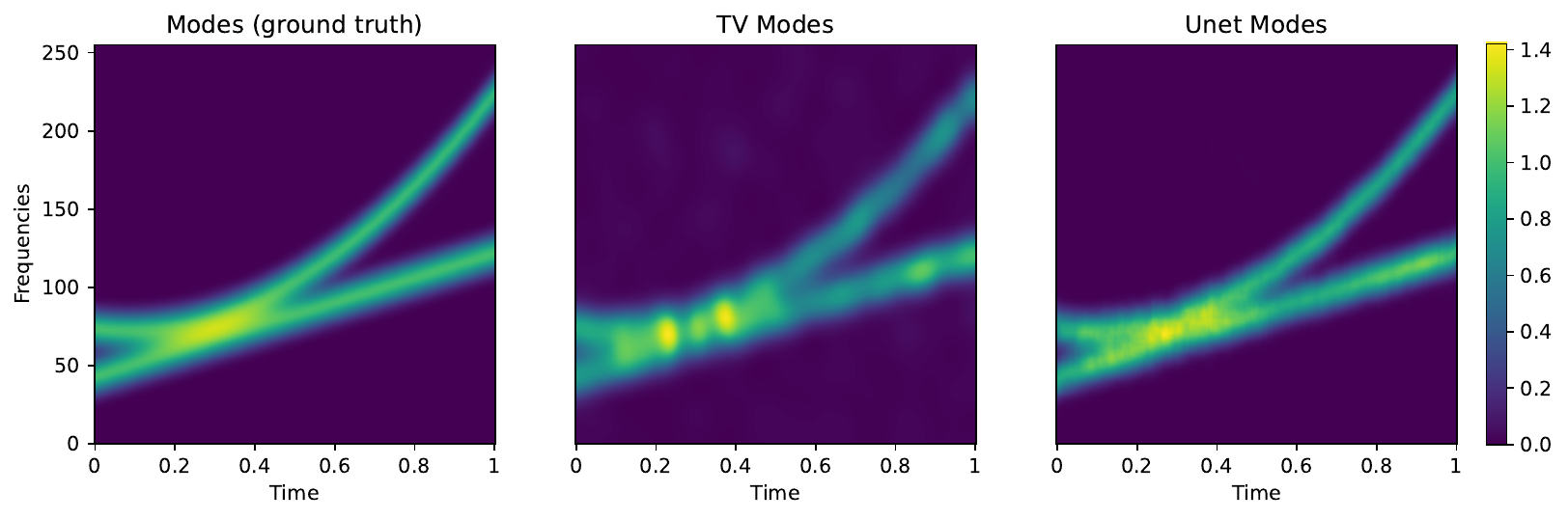}\label{figa:gaussian}
    }
    \subfloat[]{ \includegraphics[ width=0.4\textwidth]{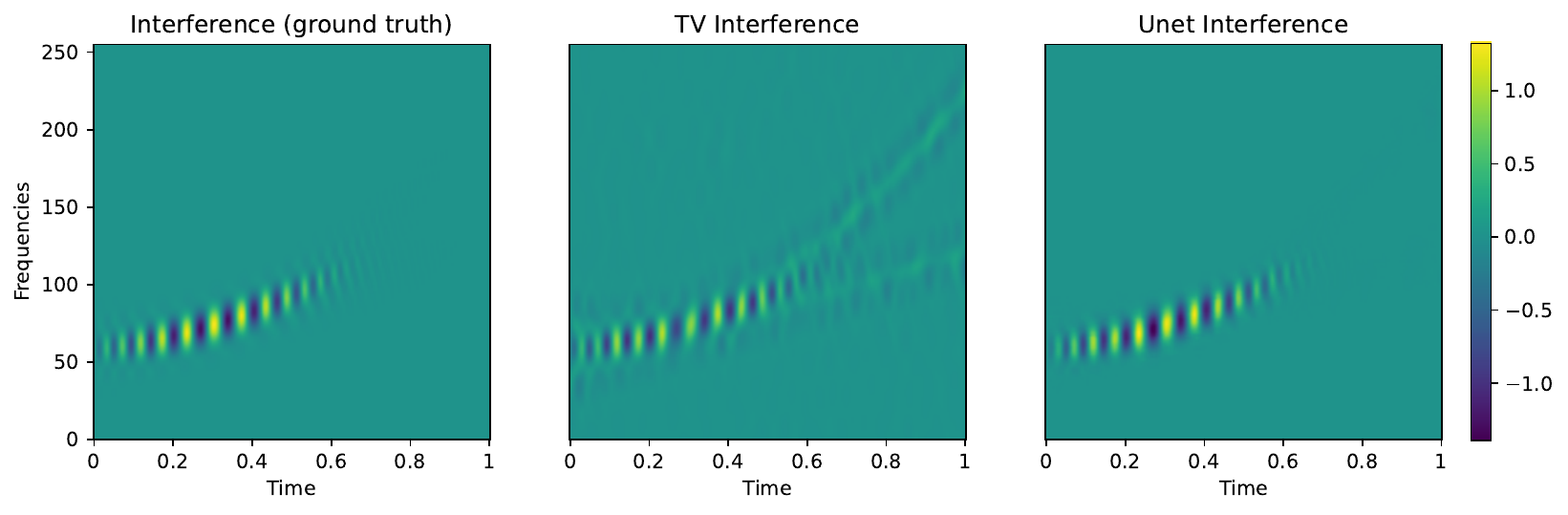}\label{figb:gaussian}
    }
    
    \subfloat[]{ \includegraphics[ width=0.145\textwidth]{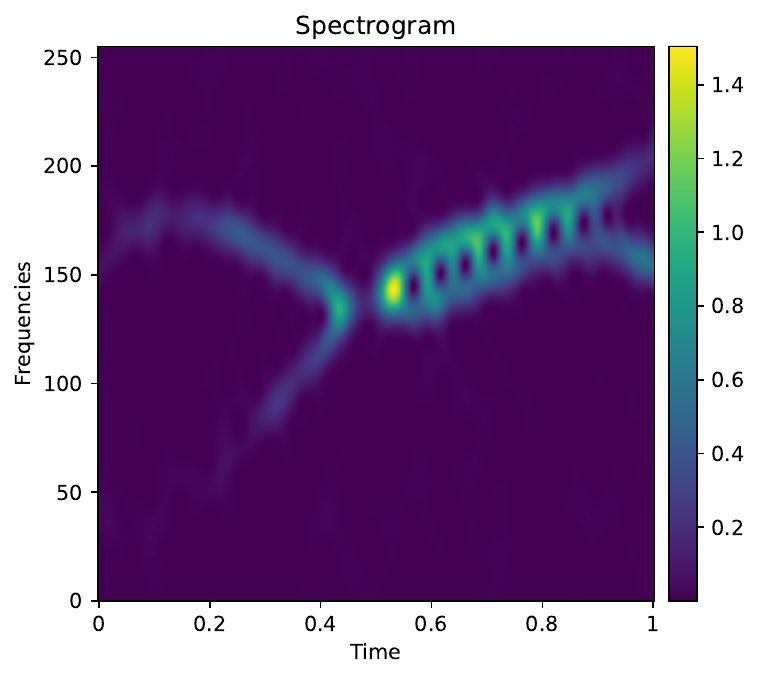}\label{figb:spect_spline}
    }
    \subfloat[]{ \includegraphics[ width=0.4\textwidth]{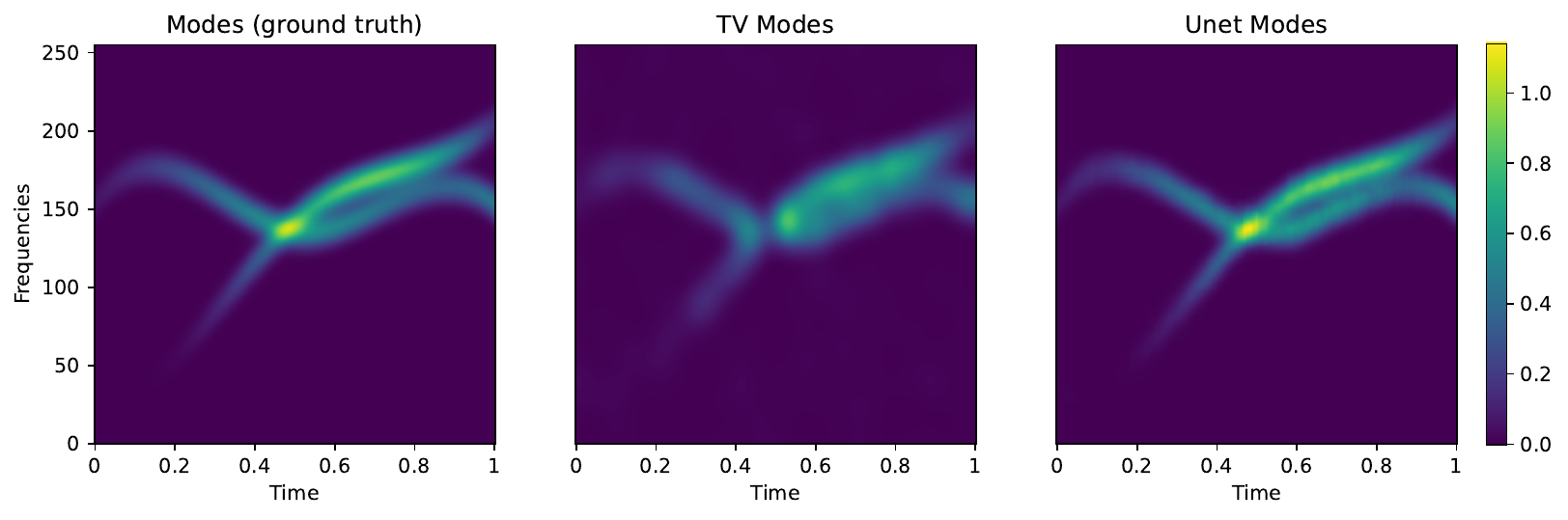}\label{figc:gaussian}
    }
    \subfloat[]{ \includegraphics[ width=0.4\textwidth]{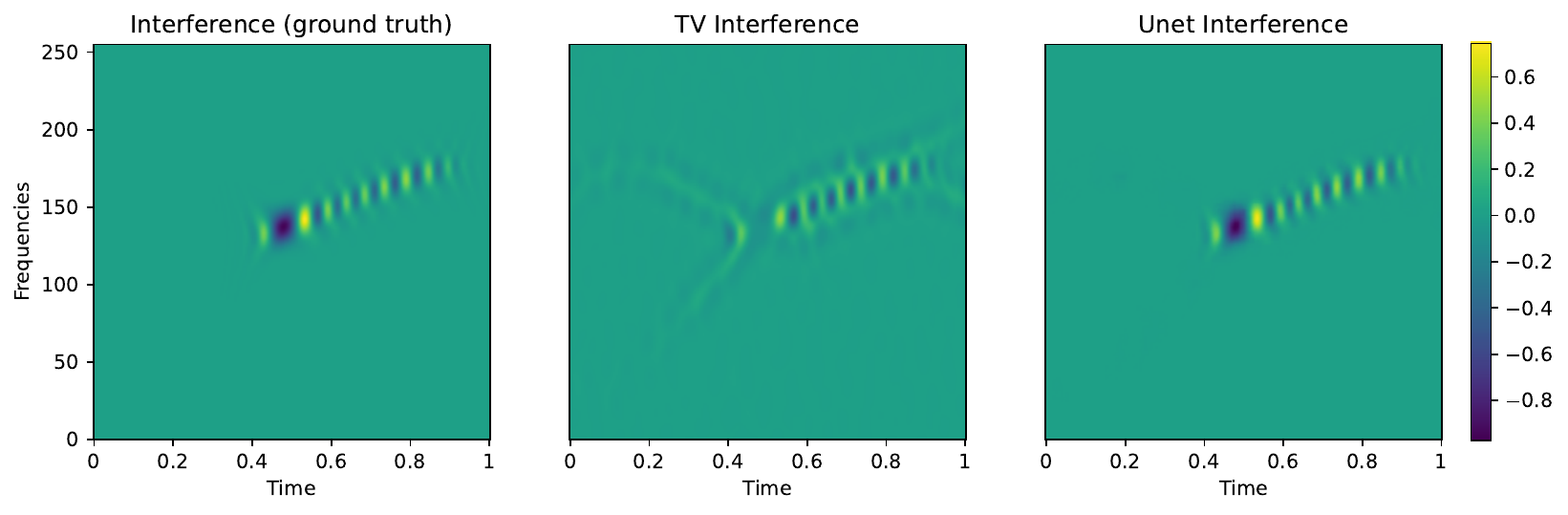}\label{fig:gaussian}
    }
    
    \caption{Spectrograms of a noisy signal consisting of (a) the superposition of a linear and hyperbolic chirps ($w=20$, $\mathrm{SNR}=10$) and (d) of two spline modes, generated by the procedure described in Section \ref{sec:synthetic} ($w=23$, $\mathrm{SNR}=10$). (b)-(e) Modes parts and (c)-(f) Interference parts in the spectrogram decomposition, presented in the following order: ground truth, TV-based decomposition, and U-Net predictions.}
    \label{fig:results}
\end{figure*}

\subsection{Synthetic Data Generation with IF and IA Splines}\label{sec:synthetic}

In our approach, we consider MCSs in which 
both the instantaneous frequencies $\phi_p'(t)$ and amplitudes $A_p(t)$ of the modes are generated using cubic splines.
For each component, we define in the TF plane 
a set of  control points $\{t_{\ell},\eta_{\ell}\}_{{\ell}=1}^{L}$, uniformly distributed in 
time and randomly sampled in frequency within a predefined range. The instantaneous frequency $\phi_p'(t)$ is thus modeled as a cubic spline function interpolating the values $\{\eta_{\ell}\}_{{\ell}=1}^{L}$ at the knots $\{t_{\ell}\}_{{\ell}=1}^{L}$, to ensure smooth variations while preventing unrealistic abrupt changes. The phase function $\phi_p(t)$ is then obtained by integration. Similarly, the amplitude function $A_p(t)$ is generated using another spline interpolation, ensuring smooth energy variations over time. We then generate $P = 2$ modes, defined as $f_p(t) = A_p(t) e^{2i\pi \phi_p(t)}$ and construct the signal  $f(t) = f_1(t) + f_2(t)$ of length $N=256$, uniformly sampled over $[0,1]$. Finally, we compute the STFT of $f$ using a Hann window $h$ of length $w$,  yielding the corresponding spectrogram image $|V_{f}^h|^2$. 
In practice, we use $ L=5 $ control points for the amplitudes and instantaneous frequencies, producing sufficient curves variability. Moreover, we ensure that the curve remains within the frequency interval $[ \Delta, N - 2\Delta ]$, with $\Delta=\frac{2N}{w}$, where $N$ is the length of the generated signal and the number of frequency-bins, so as to avoid aliasing. A Gaussian white noise $\epsilon$ is added to the signal $f$, the input SNR being randomly chosen for each image, following a uniform distribution between 0 and 20. The Hann window size $w$ is also randomly selected between 10 and 80, and the resulting spectrogram image is of size $N\times N$. For each noiseless spectrogram $\sg = |V_{f}^h|^2$ of two interfering modes, the terms \eqref{eq:t1}-\eqref{eq:t3} are computed and denoted by $\ub$ and $\vb$, respectively. The U-Net input is $\mathbf{\tilde s} = |V_{f+\epsilon}^h|^2$, representing the spectrogram of the noisy signal, while the label output $(\ub, \vb)$ consists of a two-channel ground truth containing the ideal modes spectrogram \eqref{eq:t1} and the interference term \eqref{eq:t3}. This procedure is repeated to generate a dataset of $M = 10^5$ triplet spectrogram images $\{\mathbf{\tilde s} _m, \ub_m, \vb_m \}_{m=1}^M$.

\subsection{Implementation Details}
\label{sec:implementation}
These $M = 10^5$ samples are then split into training and test datasets. The number of downsampling/upsampling levels is set to 3. Mean Squared Error (MSE) is used as loss function, and the Adam optimizer is employed with a learning rate of $5\cdot 10^{-4}$ and a weight decay of $5\cdot 10^{-8}$. Training is performed for 40 epochs, and to enhance generalization, random cropping is applied to training images as a data augmentation technique.

\section{Results and Discussions}
\label{sec:rres_discuss}

This section presents two different types of numerical experiments. We first propose a comparison between the variationnal and UNet approaches regarding mode and interference parts separation on spectrograms of MCSs. 
Secondly, we investigate how this separation can be used to estimate the IFs of 
the modes. All the experiments conducted in Python can be reproduced using the Jupyter notebook \cite{code}.
\subsection{Separating Mode parts from Interference Part} 
The comparison between the approaches introduced in Sec. \ref{sec:first_model} and \ref{sec:second_model} is made on noisy spectrogram examples with an input SNR of 10 dB, shown in Fig. \ref{fig:results} (a)-(d). Fig. \ref{figa:spec_hyperb} corresponds to the spectrogram of a signal composed of the superimposition of a linear chirp and a hyperbolic chirp, whose closely spaced modes interfere (using a Hann analysis window of size $w=20$). Fig. \ref{figb:spect_spline} represents the spectrogram of two interfering spline modes, generated using the procedure described in Section \ref{sec:synthetic} (with $w=23$), which was not included in the training or test dataset.
By construction, we have access to \emph{ground truth} (GT) images $(\ub, \vb)$ corresponding to the spectrograms of the individual modes \eqref{eq:t1} and the interference resulting from the cross terms \eqref{eq:t3}. GT images  $(\ub, \vb)$ are shown on the left of each sub-figure in Fig. \ref{fig:results} (b)-(c)-(e)-(f). The effectiveness of the two approaches in separating the mode part from the interference part is evaluated with respect to GT images.

In Algorithm \ref{Algo1} we set $\lambda=1$, $\mu=100$, $\epsilon=10$ and the number of iterations in Chambolle's algorithm to perform the projection $\pi$ is set to 100. The inputs are the spectrograms $\mathbf{\tilde s}$ in Fig. \ref{fig:results} (a)-(d), and the outputs $(\ub_{\mathrm{TV}}, \vb_{\mathrm{TV}})$ are displayed in the middle of each sub-figure of Fig. \ref{fig:results} (b)-(c)-(e)-(f), while the U-Net predictions $(\ub_{\mathrm{unet}}, \vb_{\mathrm{unet}})$ appear to their right. It can be observed that Algorithm \ref{Algo1} successfully captures the interference in the texture component (along with associated noise), but some geometric structures related to the modes remain. Moreover, the TV minimization term produces a somehow diffuse estimation of the mode part. In contrast, the U-Net predictions are closer to GT images $(\ub, \vb)$: the interference term is accurately captured and denoised, while the geometric component preserves the mode spreading, albeit with some irregularities. 
Details about the evolution of the loss function during the training and of the PSNR between 
prediction and GT images can be found in the notebook \cite{code}.

It is worth noting that the network is trained on spectrogram data generated with analysis windows of varying sizes, demonstrating its adaptability. By fixing the analysis window size, the predictions could be improved, and irregularities could be reduced. Similarly, by prioritizing training images with high-energy interference patterns, the number of required training samples could be significantly reduced, at the cost of reduced adaptability. Indeed, in many cases, the interference between the two spline modes is weak, either because they do not cross or are not sufficiently close, or because the window size is too large to induce significant interference.

\subsection{Adapting the Window Length Based on Interference Patterns}
\label{sec:refit}
\begin{figure}
\subfloat[]{ \includegraphics[width=0.23\textwidth]{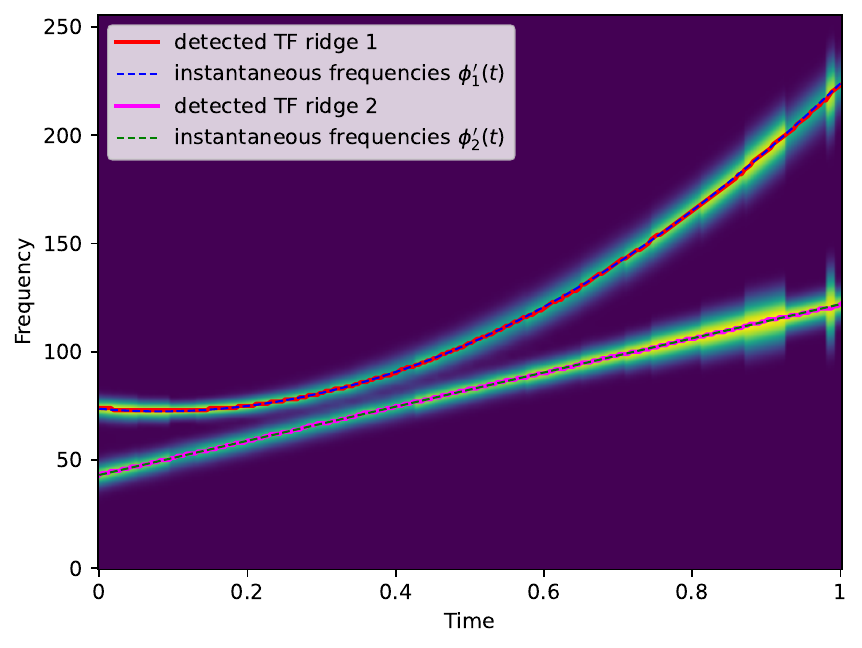}\label{figa:adapt}
}
\subfloat[]{ \includegraphics[width=0.23\textwidth]{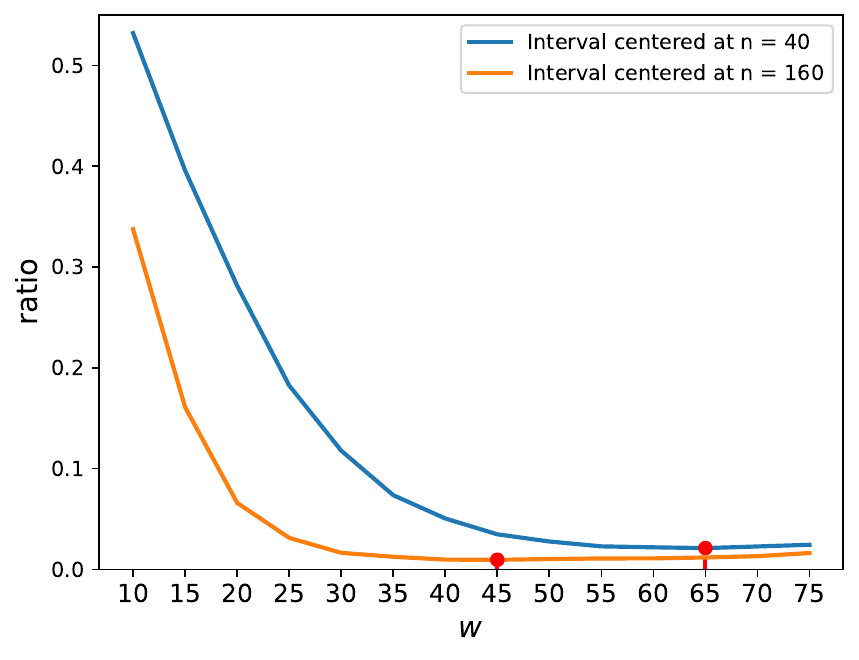}\label{figa:ratio}
}
\caption{(a) $V_{\mathrm{adap}}$ with detected TF ridges, (b) evolution of the interference-to-signal ratio as a function of window size $w$ at time indices $n=40$ and $n=160$ (resp. $t_n=0.15$ and $t_n=0.62$ on the time axis), with the minimum ratio achieved at $\hat q_n$ \eqref{eq:opt_q} highlighted in red.}
\label{V_app}
\end{figure}

In this section, we investigate how to select an appropriate 
window length based on the proposed decomposition into mode and interference components. We do not explore adaptive window selection in both time and frequency, but this could be easily achieved using the framework developed hereafter. Let $(\ub^{(w)},\vb^{(w)})$ denote the decomposition in mode and interference components obtained from the trained U-Net applied to the initial noisy spectrogram image $V_{\tilde f}^{h(w)}$, where $h(w)$ represents the Hann window of length $w$. 
Our strategy for determining an optimal window length involves first defining a collection of candidate windows, ${\boldsymbol{w}} = \{ w_q, {q=1,\dotsc,Q}\}$, 
leading, for each $q$, to different decompositions $(\ub^{(w_q)},\vb^{(w_q)})$ for each $\mathbf{\tilde s}^{(w_q)} = V_{\tilde f}^{h(w_q)}$. 
Then, for each time index $n$, we estimate locally the index $q$ associated   
with the minimum relative contribution of the interference 
to the noisy spectrogram, namely: 
\begin{eqnarray}
\label{eq:opt_q}
\hat q_n = \mathop{\rm argmin}_{q} \frac{\sum\limits_{m \in {\cal V}_n,k} |\vb^{(w_q)}[m,k]|}{\sum\limits_{m \in {\cal V}_n,k} |\mathbf{\tilde s}^{(w_q)}[m,k]|},
\end{eqnarray}
where ${\cal V}_n$ is an interval centered at $n$, whose length is determined based on the window size in $\boldsymbol{w}$.
Using this analysis, we define a novel TF representation $V_{\mathrm{adap}}$, by selecting the mode component corresponding to the optimal window $\hat q_n$ for each $n$:
\begin{eqnarray}
V_{\mathrm{adap}} [n,k] =  \ub^{(w_{\hat q_n})}[n,k], \; \forall k.
\end{eqnarray}
This new TF representation is expected to be less affected by interference than 
a single window spectrogram, enabling an improved estimation of the IFs of the modes in an MCS when they do not cross. To achieve this, RD is performed using the classical approach introduced in \cite{Carmona1999a}. 
As an illustration, we consider the two-chirp signal involved in Fig. \ref{figa:spec_hyperb} and evaluate the spectrogram for different window lengths $\boldsymbol{w}={15,20,\dotsc,75}$ at an input SNR of 10 dB. RD is then performed on $V_{\mathrm{adap}}$.
In Fig. \ref{figa:adapt}, we display the image $V_{\mathrm{adap}}$ along with the detected ridges, demonstrating how knowledge of the interference structure enables better window size selection, facilitating more accurate IF estimation. In Fig. \ref{figa:ratio}, we illustrate the process of window length selection for two different characteristic times.
\section{Conclusion and Perspectives}
In this work, we introduced two novel approaches for separating mode components from interference in time-frequency representations, leveraging both variational methods and deep learning. We compared a TV-based algorithm with a U-Net model trained on synthetic data, demonstrating the ability of both methods to extract the oscillatory interference while preserving the geometric structures of the modes.
Building on mode-interference separation, we proposed an adaptive window selection strategy that minimizes the relative contribution of interference to construct an enhanced time-frequency representation, facilitating more accurate IF estimation. 

To the best of our knowledge, applying cartoon-texture decomposition algorithms to disentangle modes from interference is a novel and original approach that warrants further investigation. Future work will focus on refining the variational framework by designing function spaces and norms better suited to the smooth, non-cartoon nature of geometric modes and the specific oscillatory patterns to be captured. Additionally, the specific noise characteristics of the spectrogram could be explicitly modeled as a third component to be separated, rather than being merged with the texture term in the current approach. Regarding the deep learning approach, alternative architectures and Plug\&Play methods will be explored to enhance separation quality and mitigate irregularities.

\bibliographystyle{IEEEtran}
\bibliography{references.bib}

\end{document}